\documentclass[letterpaper, 10 pt, conference]{ieeeconf}  
\usepackage{graphicx}
\usepackage[table,x11names]{xcolor}
\usepackage{subcaption}
\usepackage{amsfonts}
\usepackage{booktabs}
\usepackage{multirow}
\usepackage{url}
\usepackage{cite}
\usepackage{balance}

\IEEEoverridecommandlockouts                              
\title{\LARGE \bf
Retrieval-Augmented Hierarchical in-Context Reinforcement Learning and Hindsight Modular Reflections for Task Planning with LLMs
}
\author{Chuanneng Sun$^1$, Songjun Huang$^1$, Haiqiao Liu$^2$, Jie Gong$^2$, and Dario Pompili$^1$
\thanks{$^1$C. Sun, S. Huang, and D. Pompili are with the Dept. of Electrical and Computer Engineering, Rutgers University--New Brunswick, NJ, USA. Emails: {\tt\small \{chuanneng.sun, songjun.huang, pompili\}@rutgers.edu}}
\thanks{$^2$H. Liu and J. Gong are with the Dept. of Civil and Environmental Engineering, Rutgers University--New Brunswick, NJ, USA. Emails: {\tt\small \{hl1138, jg931\}@soe.rutgers.edu}
}
}

\begin{document}

\maketitle
\thispagestyle{empty}
\pagestyle{empty}

\begin{abstract}
Large Language Models~(LLMs) have demonstrated remarkable abilities in various language tasks, making them promising candidates for decision-making in robotics. Inspired by Hierarchical Reinforcement Learning~(HRL), we propose Retrieval-Augmented in-context reinforcement Learning~(RAHL), a novel framework that decomposes complex tasks into sub-tasks using an LLM-based high-level policy, in which a complex task is decomposed into sub-tasks by a high-level policy on-the-fly. The sub-tasks, defined by goals, are assigned to the low-level policy to complete.
To improve the agent's performance in multi-episode execution, we propose Hindsight Modular Reflection~(HMR), where, instead of reflecting on the full trajectory, we
let the agent reflect on shorter sub-trajectories to improve reflection efficiency. We evaluated the decision-making ability of the proposed RAHL in three benchmark environments--ALFWorld, Webshop, and HotpotQA. The results show that RAHL can achieve an improvement in performance in 9\%, 42\%, and 10\% in 5 episodes of execution in strong baselines.
Furthermore, we also implemented RAHL on the Boston Dynamics SPOT robot. The experiment shows that the robot can scan the environment, find entrances, and navigate to new rooms controlled by the LLM policy.

\end{abstract}

\section{Introduction}
The recent advent of Large Language Models~(LLMs)~\cite{brown2020language, chatgpt, chowdhery2023palm, touvron2023llama, kopf2023openassistant} has revolutionized Artificial Intelligence (AI), prompting researchers to re-examine existing algorithms and applications within the context of these powerful models. LLMs have demonstrated remarkable few-shot in-context learning capabilities through prompts~\cite{brown2020language, kojima2022large}, even surpassing traditional gradient-based approaches.
As a result, AI models built upon LLMs can be tailored to user needs without expensive fine-tuning or retraining, while still achieving competitive performance.
One particularly exciting area of research is the application of LLMs to robotic applications, including path planning~\cite{xie2023languageconditioned, zu2024language}, grasping~\cite{tziafas2023languageguided, rashid2023language}, task planning~\cite{song2023llm, rana2023sayplan, zhou2024isr, graule2024gg, chen2024autotamp, arora2024anticipate}, skill sythesis~\cite{ha2023scaling, zhang2023bootstrap}, scene understanding~\cite{majumdar2023findthis}, manipulation~\cite{ren2022leveraging, shridhar2022perceiveractor, zhou2022modularity, nair2022rm, xia2024kinematic}, etc.
In these works, LLMs serve as policies
and sometimes as evaluators.


\begin{figure*}[!t]
    \centering
    \begin{subfigure}{0.6\textwidth}
      \includegraphics[width=\linewidth]{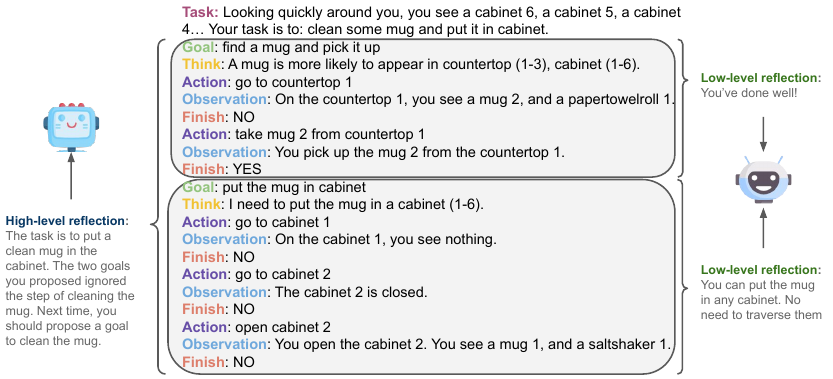}
      \caption{\label{fig:demo}}
    \end{subfigure}\hfil
    \begin{subfigure}{0.36\textwidth}
      \includegraphics[width=\linewidth]{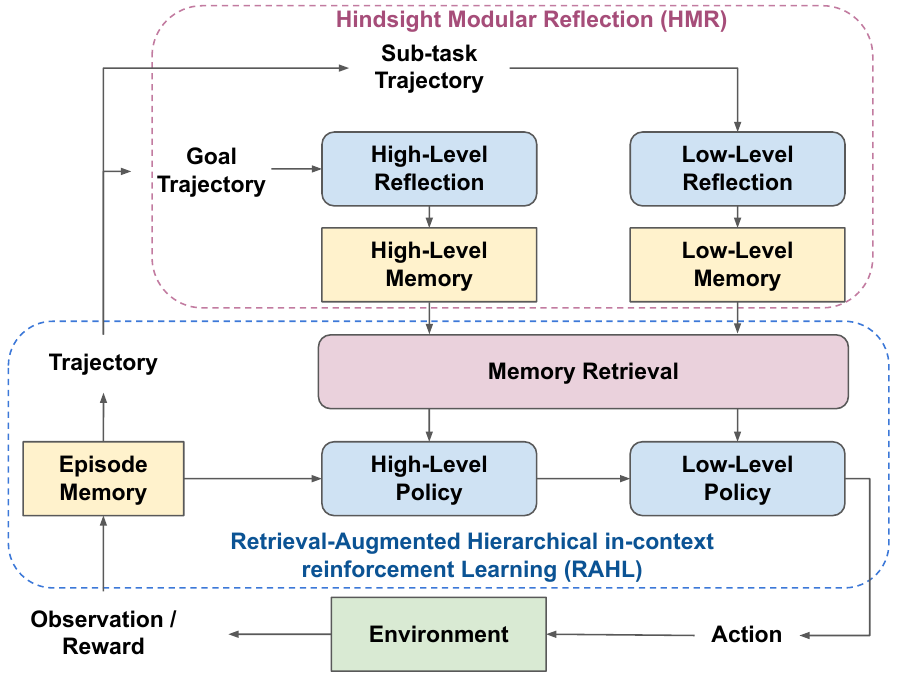}
      \caption{\label{fig:flowchart}}
    \end{subfigure}
    \caption{(a)~Demonstration of a typical task in ALFWorld. The high-level reflection attempts to correct the errors made by the high-level policy, while the low-level reflection corrects the errors from the low-level policy. (b)~Flow diagram of the proposed Retrieval-Augmented Hierarchical in-context reinforcement Learning~(RAHL) and Hindsight Modular Reflection~(HMR).
    }
    \vspace{-.1in}
\end{figure*}




In this work, we consider LLM-based task planning, where we transform LLMs into a Reinforcement Learning~(RL) policy. Although several LLM-based task planning frameworks have been proposed, most of them do not close the loop, and, as a result, there is no policy improvement over multi-episodic execution.
In this work, we propose Retrieval-Augmented Hierarchical in-context reinforcement Learning~(RAHL), a simple yet effective framework that combines Retrieval Augmented Generation~(RAG) and goal-conditioned Hierarchical Reinforcement Learning~(HRL) to enhance the performance of in-context RL. In RAHL, a high-level policy proposes sub-goals for a low-level policy to accomplish, enabling the decomposition of complex tasks into manageable sub-tasks. During this process, experiences summarized from previous sub-tasks will be retrieved from the memory as augmentation.
Furthermore, we propose Hindsight Modular Reflection (HMR) to facilitate multi-episode learning. HMR decomposes the reflection process into two components: (1)~\emph{low-level reflection}, which focuses on the actions taken to achieve each sub-goal, and (2)~\emph{high-level reflection}, which considers the overall sequence of proposed sub-goals. By providing hierarchical reflections, HMR enables the agent to identify areas for improvement and refine its strategies accordingly. 

Fig.~\ref{fig:demo} presents a visual demonstration of the proposed RAHL framework with an example from the ALFWorld environment with the task: \emph{cool some mug and put it in cabinet}. The generated action, which can be regarded as a plan because it is not directly mapped to the robot control signal, is sent to a lower-level executor to perform bottom-level control, i.e., a sequence of control signals.
Unlike traditional HRL methods, such as Hierarchical Actor Critic~(HAC)~\cite{levylearning} and option-critic~\cite{bacon2017option}, the proposed method focuses more on language-integrated learning, that is, our method transforms a general LLM to an LLM that can perform decision-making and improve itself in a multi-episode execution process.

To evaluate the effectiveness of RAHL, we conduct experiments on three diverse decision-making datasets/environments: ALFWorld~\cite{shridhar2021alfworld}, an indoor household environment for robotic tasks; Webshop~\cite{yao2022webshop}, an online shopping environment where the agent acts as a shopping assistant; and HotpotQA~\cite{yang2018hotpotqa}, a database of search-based questions and answers. Our results demonstrate that RAHL significantly outperforms existing in-context learning methods. Specifically, RAHL achieves a 9\% improvement in the success rate in ALFWorld, a 42\% improvement in Webshop (establishing a new state-of-the-art) and a 10\% improvement on HotpotQA in five episodes. Moreover, we also demonstrate the performance of RAHL using a robot experiment using Boston Dynamics SPOT.

Our \textbf{contribution} can be summarized as follows.
\begin{itemize}
    \item We propose RAHL, a decision-making framework where the high-level policy decomposes the task into sub-tasks for the low-level policy to complete.
    \item We propose HMR to promote the performance of verbal reflection. Instead of reflecting on the full trajectory, which could be inefficient because of the length of the trajectories, we propose two levels of reflection. In the low-level reflection, the low-level policy
    reflects on sub-trajectories separated by goals, while in the high-level reflection, the high-level policy reflects on the sequence of proposed goals.
    \item We perform evaluations in three different benchmark environments and show that RAHL can achieve improvements over strong baselines. We also performed an ablation study to further quantify the contribution to the performance of each part of the framework.
    \item We conducted hardware experiments to demonstrate the hardware integration capability of the proposed method. The robot used is a Boston Dynamics SPOT with two cameras mounted on the robot head facing front left and front right, respectively.
\end{itemize}

\section{Related Work}
\label{sec:related_work}

\textbf{LLM-based task planning:}
There are some works that study LLM-based task planning. However, some of these works~\cite{arora2024anticipate} directly combine LLMs with task-planning languages without further improvements. Other frameworks~\cite{song2023llm, rana2023sayplan, zhou2024isr, graule2024gg, chen2024autotamp} have more sophisticated designs, but do not incorporate environmental feedback for policy improvement. Although ISR-LLM~\cite{zhou2024isr} adopts a feedback loop to improve the LLM-generated plan before execution, the feedback is based on an LLM evaluator, which might be prone to errors.

\textbf{In-context RL:}
As opposed to gradient-based RL, in-context RL does not train the model directly, but guides the model with contexts (e.g., instruction or few-shot examples). Yao et al.~\cite{yao2023react} proposed ReAct, in which LLMs are guided to generate tractable thought traces to solve the problem. ReAct is an open-loop approach because there is no feedback involved. Building upon ReAct, Shinn et al.~\cite{shinn2024reflexion} proposed Reflexion as a closed-loop solution, in which the reflection on the past episode is generated by the LLMs and used as part of the context for the next episode. Another work, ADaPT~\cite{prasad2023adapt}, also adopts the open-loop design and generates a plan to decompose a task into smaller sub-tasks at the beginning of execution. In addition, ADaPT employs a task completion heuristic to determine whether a task is achievable. If not, they will break the task again into smaller sub-tasks.
Brook et al.~\cite{brooks2024large} proposed model-based in-context learning to implement policy iteration without training the LLM parameters. The LLMs are used to predict the next state given the current state and action and are also used as the action value function.
Hao et al.~\cite{hao2023reasoning} proposed RAP, where they regard the LLM as the world model and the decision-maker with a reasoning tree as the backbone. Murthy et al.~\cite{murthy2023rex} proposed REX, in which a Monte Carlo tree is used as the backbone to guide agent exploration combined with Upper Confidence Bound~(UCB) to balance exploration and exploitation. Zhao et al. proposed ExpeL~\cite{zhao2024expel}, which first collects a few episodes of trajectories and uses them to generate a set of rules as insights to guide future executions. Belkhale et al.~\cite{belkhale2024rt} proposed RT-H, a hierarchical planning framework for robotic manipulation. Their framework is closely integrated with the Vision Language Model~(VLM), which first predicts an action in natural language, and then translates this action into robot actions. 
Among these methods, 
RT-H and ADaPT also adopt a hierarchical structure, but there is no policy improvement after the first episode.

\section{Proposed Work}\label{sec:prop_work}

In this section, we first define the problem for our work in Sect.~\ref{sect:prop:hrl}. We then introduce the proposed RAHL in Sect.~\ref{sect:prop:RAHL}. 

\subsection{Problem Definition} \label{sect:prop:hrl}

RL algorithms can solve problems modeled as a Markov Decision Process~(MDP). An MDP is defined by a state space $\mathcal{S}$, which characterizes the system's properties, and an action space $\mathcal{A}$. The core component of an RL agent is the policy $\pi: \mathcal{S} \rightarrow \mathcal{A}$, which maps states to actions. After executing an action, the environment transitions to a new state according to a possibly unknown state transition function $\mathcal{T}: \mathcal{S} \times \mathcal{A} \rightarrow \mathcal{S}$ and generates a reward for the agent, defined by a reward function $r: \mathcal{S} \times \mathcal{A} \rightarrow \mathbb{R}$. The goal of RL algorithms is to maximize the expected return $G=\sum_{t=0}^T \gamma^t r_t$, where $T$ is the maximum number of steps, $\gamma$ is the discount factor, and $r_t$ is the reward.

In our work, we consider a retrieval-augmented goal-conditioned MDP based on the conventional MDP introduced above, where the action depends not only on the observation but also on a goal, $g \in \mathcal{G}$, generated by a high-level policy, $\pi_h: \mathcal{S}_h \rightarrow \mathcal{G}$, where $\mathcal{S}_h$ is the state space augmented with high-level memory. Given the goal, the low-level policy $\pi_l: \mathcal{S}_l \times \mathcal{G} \rightarrow \mathcal{A}$, where $\mathcal{S}_l$ is the low-level memory-augmented state space, makes decisions to achieve the goal. 
In practice, since the state space actually consists of texts, we concatenate the retrieved reflections generated by HMR with the original state as input to the LLM policy.
To determine whether a goal has been reached, we introduce the finisher, $F: \mathcal{S} \times \mathcal{G} \rightarrow \{0, 1\}$. 
The finisher determines whether a goal has been achieved by examining the execution history of the sub-task. It outputs ``Yes" if the goal is completed and ``No" otherwise. In particular, the finisher only takes the trajectory after the last completed goal as input. If the finisher determines that the current goal is achieved, the high-level policy is queried to propose a new goal to progress toward completing the main task. If the goal has not yet been achieved, the low-level policy is queried to generate another action.

\subsection{RAHL} \label{sect:prop:RAHL}

RAHL leverages the in-context learning ability of LLMs by formatting relevant information as prompts. The prompt in our work consists of three parts--\emph{i) the few-shot examples, ii) the retrieved high- and low-level reflections from the memory, and iii) the tags to guide the generation process.} In the rest of this section, we will introduce these components and the decision-making workflow. The structure of the proposed framework is shown in Fig.~\ref{fig:flowchart}. 

\textbf{Hierarchical in-context decision-making:}
The generation process can be viewed as a hierarchical decision-making process. The high-level policy in this process generates goals, whereas the low-level policy generates actions to achieve those goals. Unlike ADaPT, which devises sub-tasks in the planning phase before execution, RAHL generates goals step by step based on the task and the history of the current episode. This online goal generation allows the agent to correct errors made in previous sub-tasks. The low-level policy focuses on achieving the given goal, regardless of the main task, effectively decomposing the complex task into smaller, more manageable sub-tasks that the language agent can solve. Following the approach of ReAct, we allow the agent to ``think" before acting, making the agent's behavior more tractable and explainable. This hierarchical process is performed by tag-guided prompting.

\begin{figure}[!t]
    \centering
    \includegraphics[width=1\linewidth]{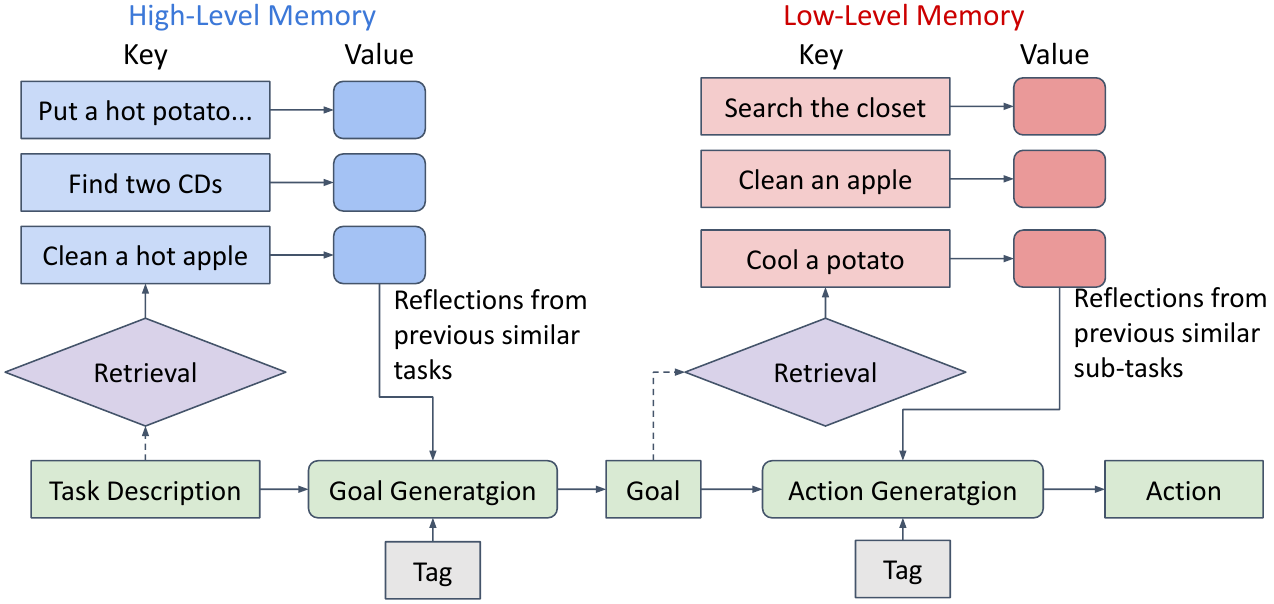}
    \caption{The decision-making process of RAHL.}
    \label{fig:generation_process}
\end{figure}

To mimic the human decision-making process, we inject prior knowledge into the process by guiding the agent with tags. Specifically, we define four kinds of tags--\texttt{[Goal]}, \texttt{[Think]}, \texttt{[Action]}, and \texttt{[Finish]}. The execution starts with a \texttt{[Goal]} tag followed by a \texttt{[Think]} tag to devise a plan for completing the goal. Then, a \texttt{[Action]} tag will be prompted to generate an action to interact with the environment. After each action, \texttt{[Finish]} will be asked to determine whether the goal is achieved. The input prompts to the LLM consist of three parts--reflections from previous trajectories, which are stored in the long-term memory; the trajectory so far, which is stored in the short-term memory; and few-shot examples. The few-shot examples contain the entire decision-making process (i.e., full trajectories) from which the agent can learn. The intuition behind using full-trajectory examples instead of modular examples is that the agent needs to consider the entire trajectory to correct its mistakes and make more logical decisions. Fig.~\ref{fig:generation_process} shows an example process of generating goals and actions. Note that goals are not updated at every step; instead, they are only updated (i.e., regenerated) when the finisher returns 1.

\textbf{Hindsight modular reflection:}
The hierarchical decision-making process might fail. As opposed to parameter-based methods, which update the model parameters when there is a failure,
we adopt modular verbal feedback to improve the agent's performance over multiple episodes, where the failed trials are summarized into one or two sentences and stored in the long-term memory for future trials. However, reflecting on the full trajectory, as adopted in Reflection, is not effective because the agent may have difficulty identifying where it went wrong due to the length of the trajectory.
To mitigate this problem, we propose HMR,
where we directly regard the goals generated by the high-level policy as the intermediate goals and encourage the low-level policy to complete this goal, even if the goal is wrong. That is, the low-level policy needs to complete the goals proposed by the high-level policy, regardless of its correctness. In other words, the action generation process can be written as,
\begin{equation}
    a = \pi_l(s_l, g) = \pi_l(s_l, \pi_h(s_h)),
\end{equation}
where $s_l$ and $s_h$ are low- and high-level reflections augmented states.

In this way, the reflection process is divided into two smaller but more specific reflection processes, corresponding to the low-level and high-level policies. The reason behind this is that shorter reflection inputs (i.e., sub-trajectories as opposed to full trajectories) lead to better reflection performances because irrelevant information is filtered. As a result, the full trajectory is decomposed into one goal trajectory, which maintains the history of proposed goals, and multiple sub-trajectories, with each one corresponding to a sequence of actions to complete a goal.

\textbf{Memory and retrieval process:}
In RAHL, we maintain a short-term memory that stores the trajectory so far in the current episode and a long-term memory that stores reflections from past experiences. Long-term memory is divided into two parts: high- and low-level memories, which store reflections from high- and low-level reflections, respectively. The long-term memory stores reflections in key-value pairs, with keys being the task or sub-task description and values being the corresponding reflections.

Once a new query comes to high- or low-level memory, the retrieval process will calculate its embeddings and subsequently calculate the distance between the query's embeddings and the keys' embeddings to find the top 2 matches. Then, the corresponding values of these 2 matches, i.e., the reflections, will be used as part of the prompt.

\section{Performance Evaluation}\label{sec:perf_eval}

In this section, we will introduce the three environments for evaluation in Sect.~\ref{sec:eval:env}, and the experiment/simulation setup and comparison plan in Sect.~\ref{sec:eval:setup}. Then we will discuss the results obtained and the results of the ablation study in Sect.~\ref{sec:eval:results} and \ref{sec:eval:ablation}, respectively.

\subsection{Environments} \label{sec:eval:env}
We conducted experiments in three environments--ALFWorld~\cite{shridhar2021alfworld}, Webshop~\cite{yao2022webshop}, and HotpotQA~\cite{yang2018hotpotqa}. These environments are not typical machine learning datasets; rather, they are interactive environments that generate observations and receive actions in the form of texts. Although these may seem to be decoupled from robotic planning, we argue that the action space in these environments can be regarded as \emph{plans}. With a plan, a low-level executor (e.g., a robot hand) can ground this plan to a sequence of control signals (e.g., move the index finger to the left by 1cm). Our work mainly focuses on the planning part, where we generate orders/commands for the robot to perform, and a low-level algorithm will translate these orders or commands to the robots' real action spaces. Therefore, the purpose of these environments is only to test the planning capacity of the proposed RAHL.

\textbf{ALFWorld:} ALFWorld is a household environment that requires an agent to make decisions over multiple steps to complete a task, such as putting a hot apple on the countertop. We follow the setup in Reflexion and run 134 scenarios across six different tasks, including moving a hidden object, cleaning an object, heating or cooling an object, etc. Typically, a task can be decomposed into several sequential sub-tasks.

\textbf{Webshop:}
Webshop is an online benchmark environment that tests the agent's ability to navigate through websites to locate items, select the correct options, and purchase items. There are two types of actions--search for an item on the search page and click a button on the page. The agent will need to extract useful information from the observation and choose the correct option according to the instructions.

\begin{figure*}[t!]
    \centering
    \begin{subfigure}{0.3\textwidth}
      \includegraphics[width=\linewidth]{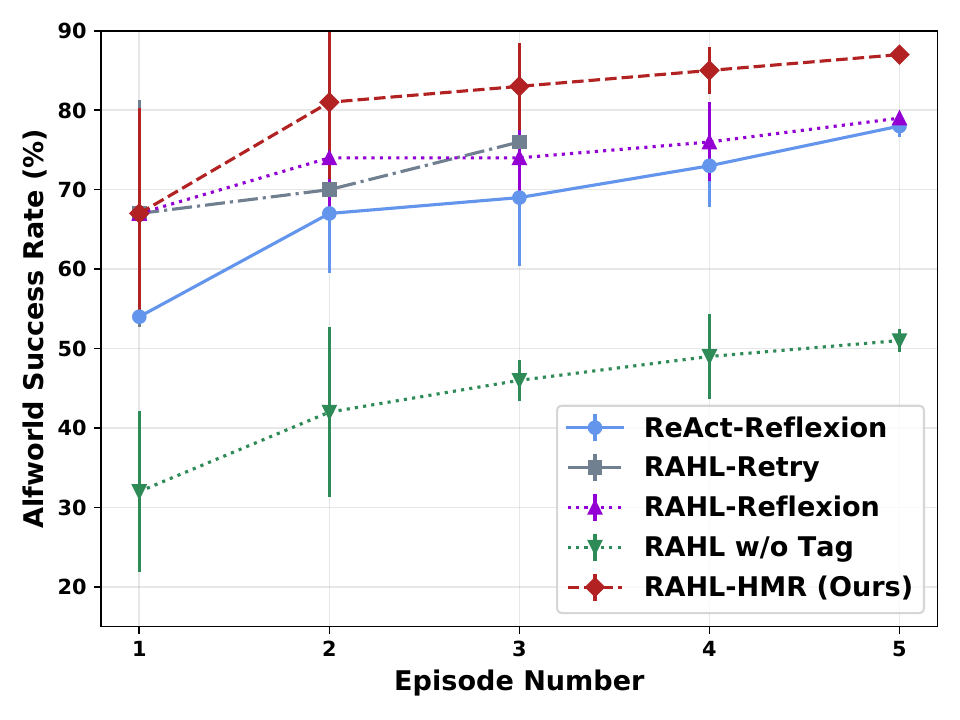}
      \caption{\label{fig:alfworld} ALFWorld success rate.}
    \end{subfigure}\hfil
    \begin{subfigure}{0.3\textwidth}
      \includegraphics[width=\linewidth]{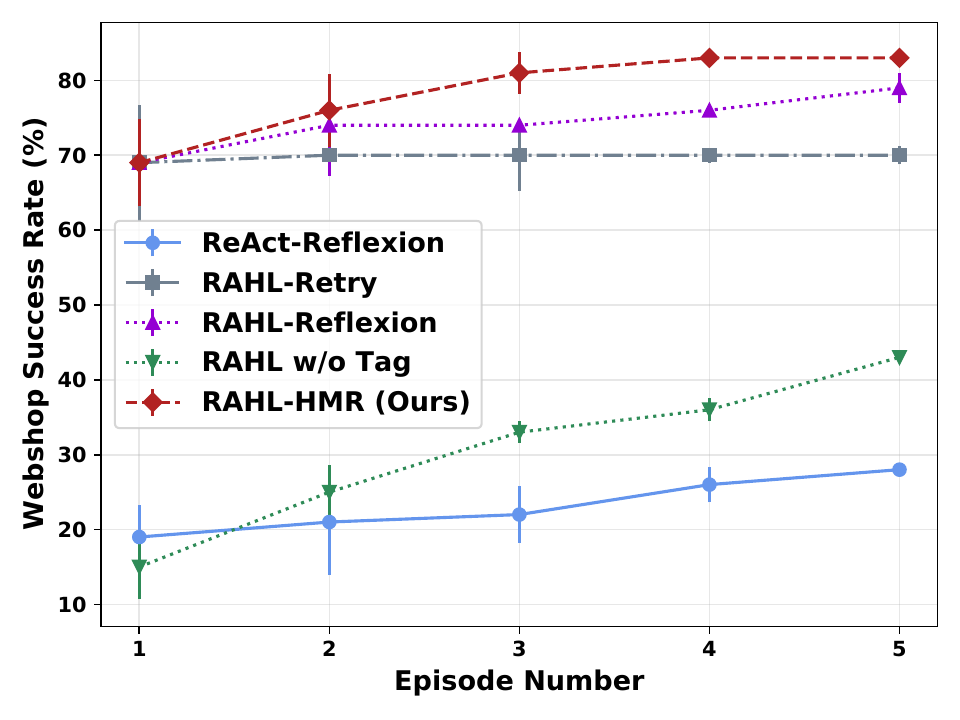}
      \caption{\label{fig:webshop} Webshop success rate.}
    \end{subfigure}\hfil
    \begin{subfigure}{0.3\textwidth}
      \includegraphics[width=\linewidth]{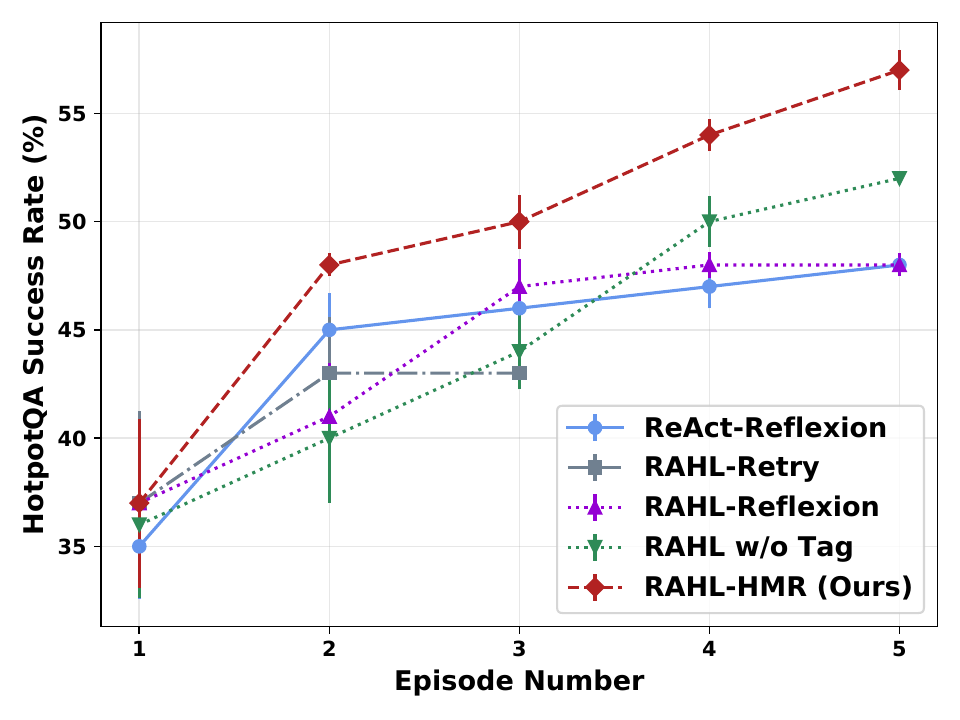}
      \caption{\label{fig:hotpotqa} HotpotQA success rate.}
    \end{subfigure}\hfil
    \caption{\label{fig:results}Success rate over five episodes in three datasets/environments using GPT-3.5-turbo. The results and confidence intervals are obtained over ten runs.
    }
\end{figure*}

\textbf{HotpotQA:} HotpotQA is a Wikipedia-based search simulator with 113K question-answer pairs. The agent needs to search, extract information from the search results, and combine information from multiple searches to obtain the final answer. There are three kinds of actions--search an entity, look up a keyword in the searched content, and finish with an answer.

\subsection{Setup and Comparison}\label{sec:eval:setup}
To show the performance of the proposed RAHL, we compare the proposed RAHL with the following frameworks.

\textbf{Retroformer}~\cite{yao2023retroformer}: Retroformer is a gradient-based framework, where the LLM is frozen and is used as the policy, while another smaller LM is trained to provide verbal feedback on the decisions based on the reward. The authors used Low Rank~(LoRA) fine-tuning~\cite{hu2021lora} to reduce the number of parameters to be fine-tuned. In our comparison, we use $rank=4$ for LoRA.

\textbf{ADaPT}~\cite{prasad2023adapt}: ADaPT will first try to solve the task, and if it fails, it will decompose the task into sub-tasks and try to solve these sub-tasks. The algorithm stops when the number of maximum recursions, $d_{max}$, is reached. Although ADaPT does not execute in multiple episodes, it tries to solve the task $d_{max}$ times. Therefore, we do not count ADaPT's results as Pass@1 results.

\textbf{Reflexion}~\cite{shinn2024reflexion}: Reflexion adopts ReAct~\cite{yao2023react} as the bottom-level actor, which can reason about the current situation and generate tractable reasoning traces. At the end of each episode, Reflexion will generate reflections for the full episode following a few reflection examples. 

\textbf{ExpeL}~\cite{zhao2024expel}: ExpeL learns from past experiences by storing them in memory and retrieving similar experiences from the memory during execution. The retrieval is based on the vector distance between the query and key embeddings. ExpeL also maintains a rule set storing the rules summarized from past experiences.

Unless otherwise specified, we use \texttt{GPT-3.5-turbo} as the LLM back-end. For Webshop and HotpotQA, we provide two full trajectories of successful execution to the agent as context. For ALFWorld, following Yao et al.~\cite{shinn2024reflexion}, we provide two task-specific examples to the agent for each of the six tasks. We also provide examples for the hindsight reflection. Specifically, we provide two examples of the goal-trajectory reflection and one example of the full-trajectory reflection.
Since ADaPT retries to execute the task when failed, we do not count its performance at \#Epi=1 as Pass@1 results.

\subsection{Results and Analysis} \label{sec:eval:results}

\begin{table}[]
\footnotesize\setlength{\tabcolsep}{3pt}%
\centering
\caption{\label{tab}Success rate in percentage for three datasets--ALFWorld, Webshop, and HotpotQA. The highlighted results represent the best results in that row. \#Epi represents the number of episodes. 
}
\begin{tabular}{c|c|ccccc}
\toprule
\textbf{Datasets} & \#Epi & Retroformer & ADaPT & Reflexion & ExpeL & RAHL \\ \midrule
\multirow{2}{*}{\textbf{Alfworld}} & 1 & 62\% & \multirow{2}{*}{71\%} & 54\% & 59\% & \cellcolor{blue!18}67\% \\
 & 5 & \cellcolor{blue!18}100\% &  & 78\% & 64\% & 87\% \\ \midrule
\multirow{2}{*}{\textbf{Webshop}} & 1 & 33\% & \multirow{2}{*}{45\%} & 19\% & 35\% & \cellcolor{blue!18}69\% \\
 & 5 & 36\% &  & 28\% & 41\% & \cellcolor{blue!18}83\% \\ \midrule
\multirow{2}{*}{\textbf{HotpotQA}} & 1 & 34\% & - & 35\% & 28\% & \cellcolor{blue!18}37\% \\
 & 5 & 53\% & - & 47\% & 39\% & \cellcolor{blue!18}57\% \\ \bottomrule
\end{tabular}
\vspace{-.1in}
\end{table}

The evaluation results of the three environments are presented in Table~\ref{tab}.
We can observe that RAHL can achieve decent Pass@1 performance (i.e., the performance at the first episode without any experiences in the past), with only ADaPT achieving a slightly better performance in ALFWorld. The reason is that ADaPT tries to approach a problem multiple times, and if it fails in the first attempt, it will decompose the task into sub-tasks and try to approach it by completing the sub-tasks. Although this seems reasonable, a complex task structure might confuse the LLM and complicate the reasoning process. Therefore, RAHL can outperform ADaPT in the Webshop.
Another observation from the results is that Retroformer can achieve decent performances when \#Epi=5, which owes to its gradient-based nature. However, since the gradient is not directly applied to the LLM agent but to the reflection agent, the Retroformer is still constrained by the ability of the reflection agent.

\begin{figure*}
    \centering
    \includegraphics[width=0.95\textwidth]{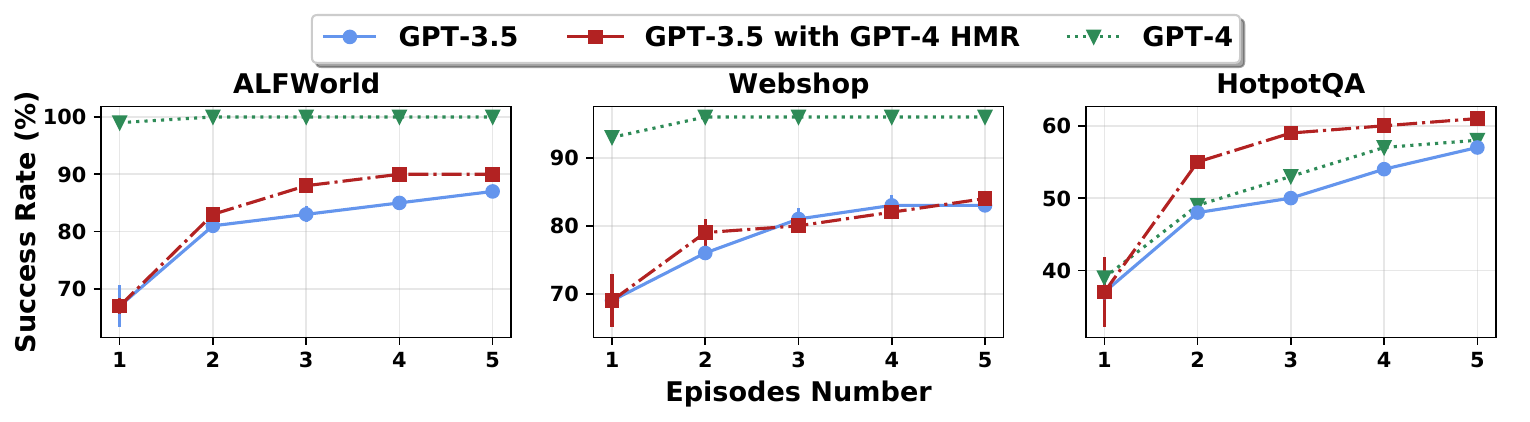}
    \caption{Success rates in percentage obtained with different LLMs. \texttt{GPT-3.5} and \texttt{GPT-4} indicate the decision-making and reflection are performed with the same type of LLMs, while \texttt{GPT-3.5 with GPT-4 HMR} indicates GPT-3.5 is used for decision-making while GPT-4 is used for reflection. The results and confidence intervals are obtained over ten runs.}
    \label{fig:gpt}
\end{figure*}

We can also observe that RAHL can outperform Reflexion in both Pass@1 and Pass@4 performances, especially in ALFWorld and Webshop, which typically have long decision-making trajectories. This demonstrates the advantage of the hierarchical structure that RAHL has. We cannot directly observe which reflection approach is better, but we will present more insight on this in Sect.~\ref{sec:eval:ablation}.


\begin{figure}[!h]
    \centering
    \begin{subfigure}{0.23\textwidth}
      \includegraphics[width=\linewidth]{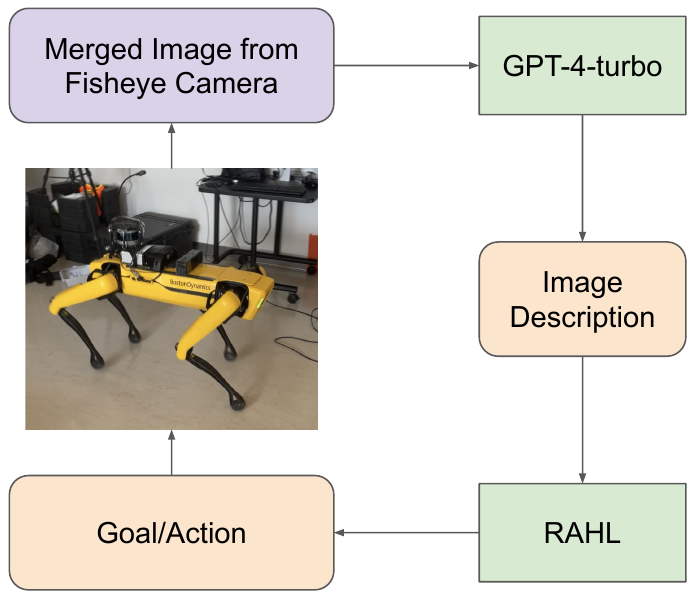}
      \caption{\label{fig:dog_diag}}
    \end{subfigure}\hfil
    \begin{subfigure}{0.25\textwidth}
      \includegraphics[width=\linewidth]{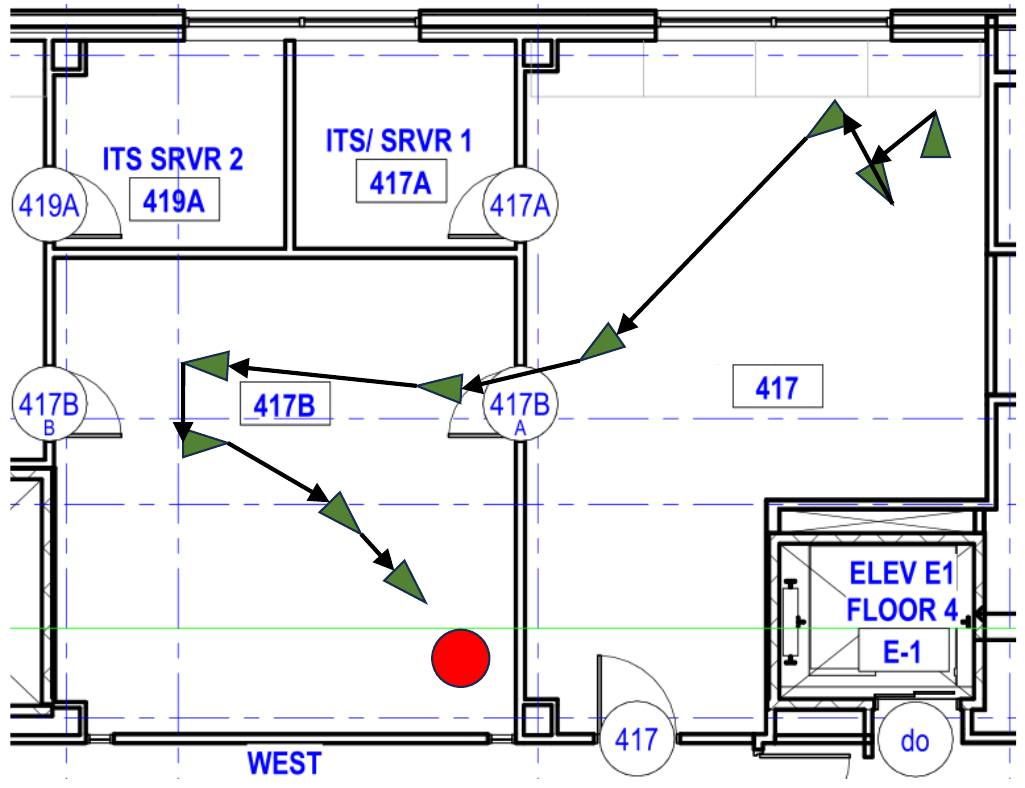}
      \caption{\label{fig:floorplan}}
    \end{subfigure}\hfil
    \caption{\label{fig:dog}(a) The diagram of the system designed for the experiment with Boston Dynamics SPOT. (b) Robot (green spearhead with the head pointing in the camera's direction) trajectory in the rooms to locate the target person (red dot).
    }
    \vspace{-.1in}
\end{figure}

\subsection{Ablation Study} \label{sec:eval:ablation}
To quantify the contribution to the performance of each part of RAHL, we performed an ablation study by comparing RAHL with three different variants of RAHL.
\begin{itemize}
    \item RAHL-Retry: We provide past failed full trajectories to the agent without any summarization or reflection. Since the trajectories are long, RAHL-Retry will terminate after a few episodes because of exceeding the maximum sequence length of the LLM.
    \item RAHL-Reflexion: We adopt Reflexion as the reflection technique where the reflection is on the full trajectory.
    \item RAHL w/o Tag: We let the LLM decide what the next step should be instead of guiding it with tags. In this way, in addition to generating actions in the environment's action spaces, it can also generate three types of actions: think, propose a goal, and finish a goal. Consequently, the reflection examples have been redesigned to remove the tags.
\end{itemize}

Note that in the experiment, RAHL-Retry, RAHL-Reflexion, and RAHL-HMR have the same first episode because they all have RAHL as the decision maker, and the difference among them is the reflection technique. After we obtain the trajectory of the first episode using RAHL, we start directly from the second episode but with different long-term memories.

The results are presented in Fig.~\ref{fig:results}. We can observe that the RAHL w/o Tag does not achieve comparable performances with other tag-guided methods, indicating that injecting the prior knowledge of humans' thought processes is helpful for LLM agents' decision-making. Moreover, RAHL-Reflexion and RAHL-HMR can outperform RAHL-Retry because experiences in human language can be perceived better by the LLM agent than plain trajectories, not to mention that retry-based methods are not scalable because of the possibility of exceeding the maximum sequence length of the LLMs. Another notable observation is that RAHL-HMR can outperform RAHL-Reflexion, because modular reflections are more efficient in identifying errors agents made, especially in long trajectories and text-heavy environments like ALFWorld (Fig.~\ref{fig:alfworld}) and HotpotQA (Fig.~\ref{fig:hotpotqa}).

We also present the results obtained using different LLM backends in Fig.~\ref{fig:gpt}. It can be observed that GPT-4 can indeed bring about a significant performance improvement for simpler environments such as ALFWorld and Webshop. The reason for this improvement is that GPT-4 has better knowledge about the world because it is trained on more and newer data than GPT-3.5. In other words, tasks in ALFWorld, a household environment, and Webshop, an online shopping environment, can be completed with general human skills and knowledge, while questions in HotpotQA require the agent to search for the information. Furthermore, we can observe that \texttt{GPT-3.5 with GPT-4 HMR} can outperform \texttt{GPT-3.5}, which further demonstrates the importance of reflection. 
Another interesting finding is that
\texttt{GPT-3.5 with GPT-4 HMR} can outperform \texttt{GPT-4} in HotpotQA, indicating that better LLMs do not necessarily lead to better performances in decision-making tasks.


\section{Hardware Experiment}
Besides the results obtained on the three benchmark environments, we also performed experiments with the Boston Dynamics SPOT robot dog. The main focus of this experiment is to show the ability of the proposed RAHL framework to be implemented in reality. The task in the experiment is navigation, where the robot needs to navigate to a person in the room next to it. To go to the destination room, the robot needs to look for a door and then look for the person after going through the door.
The system used in the experiment and the picture of the SPOT robot are shown in Fig.~\ref{fig:dog_diag}.

We design two experiments. In the first test, the robot faces the door in the beginning. The goal of this trial is to test the compatibility of GPT-4-turbo as the image analyst and GPT-4o as the decision-maker. As a result, the robot was able to recognize the door and go through the door to look for the person.
The second test increases the difficulty by positioning the robot in a corner of the room. To reach the person, one needs to scan the first room to find the door and then go through the door to find the person. The trial focuses more on the ability of the GPT-4o as the decision-maker. The robot was successful in finding the person in the end after exploring the first room. The trajectory of the second trial is presented in Fig.~\ref{fig:floorplan}.

\balance

\section{Limitation and Discussion}\label{sec:discussion}
The proposed RAHL is essentially a deterministic policy for decision-making, with HMR being the update of the policy gradient. Though it can achieve decent performance and enhance interpretability, it still underperforms trained large models and has performance bottlenecks
in complex tasks such as HotpotQA.
The reason for this is that LLMs rely heavily on the knowledge base acquired during their training on a large human language corpus. However, this knowledge base might be biased, causing LLMs to be stubborn in certain scenarios. This stubbornness is deep rooted in the knowledge of LLM. To tackle this, we will work on better reflection techniques or methods beyond reflections in the future to conquer this issue.

\section{Conclusion}\label{sec:conclusion}
In this paper, we introduced Retrieval-Augmented in-context reinforcement Learning~(RAHL), an in-context learning framework that decomposes complex tasks into simpler sub-tasks. A high-level policy is responsible for proposing goals that define sub-tasks, while a low-level policy makes decisions to achieve the goals. To enable the framework to improve over multiple episodes, we proposed Hindsight Modular Reflection~(HMR), where we introduced low-level reflection and high-level reflection. The proposed framework was evaluated in three environments and the results showed that RAHL can outperform existing frameworks in both Pass@1 performances and multi-episode performances.
In the future, we plan to embed RAHL on drones as a planner, with a trained executor to map the plans to robot control signals and evaluate the system's performance as a whole.

\section*{Acknowledgement}
This work is supported by the NSF RTML Award No. CCF-1937403.

\bibliographystyle{ieeetr}
\bibliography{refs, survey}

\begin{thebibliography}{10}

\bibitem{brown2020language}
T.~Brown, B.~Mann, N.~Ryder, M.~Subbiah, J.~D. Kaplan, P.~Dhariwal,
  A.~Neelakantan, P.~Shyam, G.~Sastry, A.~Askell, {\em et~al.}, ``Language
  models are few-shot learners,'' {\em Advances in neural information
  processing systems}, vol.~33, pp.~1877--1901, 2020.

\bibitem{chatgpt}
OpenAI, ``Chatgpt: Large language model trained by openai.''
  \url{https://openai.com/blog/chat-gpt-3/}, 2021.

\bibitem{chowdhery2023palm}
A.~Chowdhery, S.~Narang, J.~Devlin, M.~Bosma, G.~Mishra, A.~Roberts, P.~Barham,
  H.~W. Chung, C.~Sutton, S.~Gehrmann, {\em et~al.}, ``Palm: Scaling language
  modeling with pathways,'' {\em Journal of Machine Learning Research},
  vol.~24, no.~240, pp.~1--113, 2023.

\bibitem{touvron2023llama}
H.~Touvron, L.~Martin, K.~Stone, P.~Albert, A.~Almahairi, Y.~Babaei,
  N.~Bashlykov, S.~Batra, P.~Bhargava, S.~Bhosale, {\em et~al.}, ``Llama 2:
  Open foundation and fine-tuned chat models,'' {\em arXiv preprint
  arXiv:2307.09288}, 2023.

\bibitem{kopf2023openassistant}
A.~K{\"o}pf, Y.~Kilcher, D.~von R{\"u}tte, S.~Anagnostidis, Z.~R. Tam,
  K.~Stevens, A.~Barhoum, D.~Nguyen, O.~Stanley, R.~Nagyfi, {\em et~al.},
  ``Openassistant conversations-democratizing large language model alignment,''
  {\em Advances in Neural Information Processing Systems}, vol.~36, 2024.

\bibitem{kojima2022large}
T.~Kojima, S.~S. Gu, M.~Reid, Y.~Matsuo, and Y.~Iwasawa, ``Large language
  models are zero-shot reasoners,'' {\em Advances in neural information
  processing systems}, vol.~35, pp.~22199--22213, 2022.

\bibitem{xie2023languageconditioned}
A.~Xie, Y.~Lee, P.~Abbeel, and S.~James, ``Language-conditioned path
  planning,'' in {\em 7th Annual Conference on Robot Learning}, 2023.

\bibitem{zu2024language}
W.~Zu, W.~Song, R.~Chen, Z.~Guo, F.~Sun, Z.~Tian, W.~Pan, and J.~Wang,
  ``Language and sketching: An llm-driven interactive multimodal multitask
  robot navigation framework,'' in {\em 2024 IEEE International Conference on
  Robotics and Automation (ICRA)}, pp.~1019--1025, IEEE, 2024.

\bibitem{tziafas2023languageguided}
G.~Tziafas, Y.~XU, A.~Goel, M.~Kasaei, Z.~Li, and H.~Kasaei, ``Language-guided
  robot grasping: {CLIP}-based referring grasp synthesis in clutter,'' in {\em
  7th Annual Conference on Robot Learning}, 2023.

\bibitem{rashid2023language}
A.~Rashid, S.~Sharma, C.~M. Kim, J.~Kerr, L.~Y. Chen, A.~Kanazawa, and
  K.~Goldberg, ``Language embedded radiance fields for zero-shot task-oriented
  grasping,'' in {\em 7th Annual Conference on Robot Learning}, 2023.

\bibitem{song2023llm}
C.~H. Song, J.~Wu, C.~Washington, B.~M. Sadler, W.-L. Chao, and Y.~Su,
  ``Llm-planner: Few-shot grounded planning for embodied agents with large
  language models,'' in {\em Proceedings of the IEEE/CVF International
  Conference on Computer Vision}, pp.~2998--3009, 2023.

\bibitem{rana2023sayplan}
K.~Rana, J.~Haviland, S.~Garg, J.~Abou-Chakra, I.~Reid, and N.~Suenderhauf,
  ``Sayplan: Grounding large language models using 3d scene graphs for scalable
  robot task planning,'' in {\em 7th Annual Conference on Robot Learning},
  2023.

\bibitem{zhou2024isr}
Z.~Zhou, J.~Song, K.~Yao, Z.~Shu, and L.~Ma, ``Isr-llm: Iterative self-refined
  large language model for long-horizon sequential task planning,'' in {\em
  2024 IEEE International Conference on Robotics and Automation (ICRA)},
  pp.~2081--2088, IEEE, 2024.

\bibitem{graule2024gg}
M.~A. Graule and V.~Isler, ``Gg-llm: Geometrically grounding large language
  models for zero-shot human activity forecasting in human-aware task
  planning,'' in {\em 2024 IEEE International Conference on Robotics and
  Automation (ICRA)}, pp.~568--574, IEEE, 2024.

\bibitem{chen2024autotamp}
Y.~Chen, J.~Arkin, C.~Dawson, Y.~Zhang, N.~Roy, and C.~Fan, ``Autotamp:
  Autoregressive task and motion planning with llms as translators and
  checkers,'' in {\em 2024 IEEE International Conference on Robotics and
  Automation (ICRA)}, pp.~6695--6702, IEEE, 2024.

\bibitem{arora2024anticipate}
R.~Arora, S.~Singh, K.~Swaminathan, A.~Datta, S.~Banerjee, B.~Bhowmick, K.~M.
  Jatavallabhula, M.~Sridharan, and M.~Krishna, ``Anticipate \& act:
  Integrating llms and classical planning for efficient task execution in
  household environments,'' in {\em International Conference on Robotics and
  Automation}, 2024.

\bibitem{ha2023scaling}
H.~Ha, P.~Florence, and S.~Song, ``Scaling up and distilling down:
  Language-guided robot skill acquisition,'' in {\em 7th Annual Conference on
  Robot Learning}, 2023.

\bibitem{zhang2023bootstrap}
J.~Zhang, J.~Zhang, K.~Pertsch, Z.~Liu, X.~Ren, M.~Chang, S.-H. Sun, and J.~J.
  Lim, ``Bootstrap your own skills: Learning to solve new tasks with large
  language model guidance,'' in {\em 7th Annual Conference on Robot Learning},
  2023.

\bibitem{majumdar2023findthis}
A.~Majumdar, F.~Xia, D.~Batra, L.~Guibas, {\em et~al.}, ``Findthis:
  Language-driven object disambiguation in indoor environments,'' in {\em 7th
  Annual Conference on Robot Learning}, 2023.

\bibitem{ren2022leveraging}
A.~Z. Ren, B.~Govil, T.-Y. Yang, K.~R. Narasimhan, and A.~Majumdar,
  ``Leveraging language for accelerated learning of tool manipulation,'' in
  {\em 6th Annual Conference on Robot Learning}, 2022.

\bibitem{shridhar2022perceiveractor}
M.~Shridhar, L.~Manuelli, and D.~Fox, ``Perceiver-actor: A multi-task
  transformer for robotic manipulation,'' in {\em 6th Annual Conference on
  Robot Learning}, 2022.

\bibitem{zhou2022modularity}
Y.~Zhou, S.~Sonawani, M.~Phielipp, S.~Stepputtis, and H.~Amor, ``Modularity
  through attention: Efficient training and transfer of language-conditioned
  policies for robot manipulation,'' in {\em 6th Annual Conference on Robot
  Learning}, 2022.

\bibitem{nair2022rm}
S.~Nair, A.~Rajeswaran, V.~Kumar, C.~Finn, and A.~Gupta, ``R3m: A universal
  visual representation for robot manipulation,'' in {\em 6th Annual Conference
  on Robot Learning}, 2022.

\bibitem{xia2024kinematic}
W.~Xia, D.~Wang, X.~Pang, Z.~Wang, B.~Zhao, D.~Hu, and X.~Li, ``Kinematic-aware
  prompting for generalizable articulated object manipulation with llms,'' in
  {\em 2024 IEEE International Conference on Robotics and Automation (ICRA)},
  pp.~2073--2080, IEEE, 2024.

\bibitem{levylearning}
A.~Levy, G.~Konidaris, R.~Platt, and K.~Saenko, ``Learning multi-level
  hierarchies with hindsight,'' in {\em International Conference on Learning
  Representations}.

\bibitem{bacon2017option}
P.-L. Bacon, J.~Harb, and D.~Precup, ``The option-critic architecture,'' in
  {\em Proceedings of the AAAI conference on artificial intelligence}, vol.~31,
  2017.

\bibitem{shridhar2021alfworld}
M.~Shridhar, X.~Yuan, M.-A. Cote, Y.~Bisk, A.~Trischler, and M.~Hausknecht,
  ``{ALFW}orld: Aligning text and embodied environments for interactive
  learning,'' in {\em International Conference on Learning Representations},
  2021.

\bibitem{yao2022webshop}
S.~Yao, H.~Chen, J.~Yang, and K.~Narasimhan, ``Webshop: Towards scalable
  real-world web interaction with grounded language agents,'' {\em Advances in
  Neural Information Processing Systems}, vol.~35, pp.~20744--20757, 2022.

\bibitem{yang2018hotpotqa}
Z.~Yang, P.~Qi, S.~Zhang, Y.~Bengio, W.~Cohen, R.~Salakhutdinov, and C.~D.
  Manning, ``Hotpotqa: A dataset for diverse, explainable multi-hop question
  answering,'' in {\em Proceedings of the 2018 Conference on Empirical Methods
  in Natural Language Processing}, pp.~2369--2380, 2018.

\bibitem{yao2023react}
S.~Yao, J.~Zhao, D.~Yu, N.~Du, I.~Shafran, K.~R. Narasimhan, and Y.~Cao,
  ``React: Synergizing reasoning and acting in language models,'' in {\em The
  Eleventh International Conference on Learning Representations}, 2023.

\bibitem{shinn2024reflexion}
N.~Shinn, F.~Cassano, A.~Gopinath, K.~Narasimhan, and S.~Yao, ``Reflexion:
  Language agents with verbal reinforcement learning,'' {\em Advances in Neural
  Information Processing Systems}, vol.~36, 2024.

\bibitem{prasad2023adapt}
A.~Prasad, A.~Koller, M.~Hartmann, P.~Clark, A.~Sabharwal, M.~Bansal, and
  T.~Khot, ``Adapt: As-needed decomposition and planning with language
  models,'' in {\em Findings of the Association for Computational Linguistics:
  NAACL 2024}, pp.~4226--4252, 2024.

\bibitem{brooks2024large}
E.~Brooks, L.~Walls, R.~L. Lewis, and S.~Singh, ``Large language models can
  implement policy iteration,'' {\em Advances in Neural Information Processing
  Systems}, vol.~36, 2024.

\bibitem{hao2023reasoning}
S.~Hao, Y.~Gu, H.~Ma, J.~Hong, Z.~Wang, D.~Z. Wang, and Z.~Hu, ``Reasoning with
  language model is planning with world model,'' in {\em NeurIPS 2023 Workshop
  on Generalization in Planning}, 2023.

\bibitem{murthy2023rex}
R.~Murthy, S.~Heinecke, J.~C. Niebles, Z.~Liu, L.~Xue, W.~Yao, Y.~Feng,
  Z.~Chen, A.~Gokul, D.~Arpit, {\em et~al.}, ``Rex: Rapid exploration and
  exploitation for ai agents,'' {\em arXiv preprint arXiv:2307.08962}, 2023.

\bibitem{zhao2024expel}
A.~Zhao, D.~Huang, Q.~Xu, M.~Lin, Y.-J. Liu, and G.~Huang, ``Expel: Llm agents
  are experiential learners,'' in {\em Proceedings of the AAAI Conference on
  Artificial Intelligence}, vol.~38, pp.~19632--19642, 2024.

\bibitem{belkhale2024rt}
S.~Belkhale, T.~Ding, T.~Xiao, P.~Sermanet, Q.~Vuong, J.~Tompson, Y.~Chebotar,
  D.~Dwibedi, and D.~Sadigh, ``Rt-h: Action hierarchies using language,'' {\em
  arXiv preprint arXiv:2403.01823}, 2024.

\bibitem{yao2023retroformer}
W.~Yao, S.~Heinecke, J.~C. Niebles, Z.~Liu, Y.~Feng, L.~Xue, R.~Rithesh,
  Z.~Chen, J.~Zhang, D.~Arpit, {\em et~al.}, ``Retroformer: Retrospective large
  language agents with policy gradient optimization,'' in {\em The Twelfth
  International Conference on Learning Representations}, 2023.

\bibitem{hu2021lora}
E.~J. Hu, P.~Wallis, Z.~Allen-Zhu, Y.~Li, S.~Wang, L.~Wang, W.~Chen, {\em
  et~al.}, ``Lora: Low-rank adaptation of large language models,'' in {\em
  International Conference on Learning Representations}, 2021.

\end{thebibliography}

\end{document}